# A Novel Task-Oriented Text Corpus in Silent Speech Recognition and its Natural Language Generation Construction Method

Dong Cao
DeepBlue Technology (Shanghai) Co., Ltd.
Addr: No.369,Weining Road,
Changning District, Shanghai, China
DeepBlue Academy of Sciences
DeepBlue Institute, Chongqing, China
caodong@deepblueai.com

Dongdong Zhang[*]
DeepBlue Academy of Sciences
DeepBlue Technology (Shanghai)Co., Ltd.
Addr: No.369,Weining Road,
Changning District, Shanghai, China
zhangdongdong@deepblueai.com

HaiBo Chen[*]
DeepBlue Technology (Shanghai)Co., Ltd.
Addr: No.369,Weining Road,
Changning District, Shanghai, China
chenhaibo@deepblueai.com

## ABSTRACT
Millions of people with severe speech disorders around the world may regain their communication capabilities through techniques of silent speech recognition (SSR). Using electroencephalography (EEG) as a biomarker for speech decoding has been popular for SSR. However, the lack of SSR text corpus has impeded the development of this technique. Here, we construct a novel task-oriented text corpus, which is utilized in the field of SSR. In the process of construction, we propose a task-oriented hybrid construction method based on natural language generation algorithm. The algorithm focuses on the strategy of data-to-text generation, and has two advantages including linguistic quality and high diversity. These two advantages use template-based method and deep neural networks respectively. In an SSR experiment with the generated text corpus, analysis results show that the performance of our hybrid construction method outperforms the pure method such as template-based natural language generation or neural natural language generation models.

## 1. INTRODUCTION
Acoustic speech recognition (ASR) has already become one of the most successful techniques in the era of artificial intelligence (AI) [1], [2]. However, millions of people around the world who lost their abilities to speak cannot enjoy the convenience of ASR [3]. Under certain circumstances such as large background noise or high confidential communications, ASR may lose its advantage or even doesn't work [4].

One possible solution is the silent speech recognition (SSR) which is an alternative for these situations [5], [6]. Many biological indicators has been used as the input for the SSR among which electroencephalography (EEG) is the most promising one [7]–[11]. EEG is a measure of electrical potentials between different locations on the skull. Krishna et.al showed that using EEG can help ASR systems overcome performance loss in the presence of noise [12]. Some studies have also reported that it is possible to decipher imagined speech directly through EEG [13]. Although a huge progress has been made in EEG-based SSR, it is impossible to evaluated the performances of different systems due to the lack of a consensus text corpus for SSR.

In the process of constructing EEG-based SSR text data sets, various NLP methods play an important role. Many methods for attention and neural sequence-to-sequence framework have been proposed [14]-[19]. A method has been proposed that of word-based and character-based sequence-to-sequence models[20]. This model is utilized for data-to-text natural language generation.

BERT has been successfully applied to a variety of natural language processing tasks such as name entity recognition and machine reading comprehensions[21]-[25]. The literature [26] first proposed to extend the BERT to the sequence generation task, and constructed a new natural language generation model based on the pre-training language model. In order to take full advantage of the BERT's context modeling capabilities, the proposed method uses a two-stage decoding process.

## 2. TEXT CORPUS FOR SSR
In the field of general speech recognition, there are a large number of mature text corpus available. These texts are organized and complete. The content covers all aspects and application fields of daily conversation. Based on these extensive text corpus, it can be further constructed a rich speech training set for general speech recognition.

Different from the general speech recognition application, the design and construction of training corpus is a challenging task in the field of SSR, because Using EEG as a biomarker for speech decoding has been popular for SSR. We want to build an SSR system based on EEG. Then, there is no ready-made 'EEG-text pair' data set that can be used for model training. We must build such a data set almost from zero. Quantitatively, if you want to build an available system similar to general speech recognition, then the amount of data required must be cumulative tens of thousands of hours of 'EEG-text pair' data sets. For most academic teams, this is a daunting task that requires a lot of manpower, material and time.

## 3. HYBRID MODELS FOR TEXT CORPUS
### 3.1 Task-Oriented Text Corpus
In order to solve the above dilemma, we can choose a compromise approach, considering that in application, the essential needs of SSR are those patients who lose their throat vocal function due to illness. Their outstanding scenario for SSR applications is life support. Therefore, based on this consideration, we avoid the need for a massive 'EEG-text pair' data set from the generic scenario

SSR. We start with the help of this task-oriented scene to construct the task-oriented text corpus. In this way, the demand for data set size is significantly reduced. Therefore, we need to construct session texts that serve the life support class in a targeted manner. Such texts will be further applied to the EEG signal data collection process to construct a task-oriented 'EEG-text pair' data set.

The characteristics of the life-assisted task-oriented session text is that the task-oriented session has a clear purpose relative to the open domain session, requiring high language quality and clear structure. Moreover, with the language habits and language styles of different users, the language presents richness and diversity in a task-oriented session. The special requirements of this kind of task-oriented session text construction bring new challenges to the design and construction of our data set.

we adopt a task-oriented corpus hybrid construction method based on natural language generation algorithm. The algorithm focuses on the strategy of data-to-text generation[20], and has the following two advantages, the first one is linguistic quality based on template-based method and the second is high diversity based on deep neural networks.

### 3.2 Task-Oriented Hybrid Models

The literature [20] based on the neural network method realizes the data-to-text natural langual generation. Inspired by this method, we construct a task-oriented hybrid construction method based on natural language generation algorithm. This method is used to generate task-oriented text corpus in the field of SSR. The algorithm includes two aspects, first constructing the seed text corpus, then generate task-oriented text corpus based on the seed text corpus using the deep neural network method.

The task-oriented hybrid models are used to construct the corpus. Specific steps are as follows.

**Step 1.** According to the requirements applied to the life aid class, design the corresponding cycle template, the template needs to be concise and regular, and can basically cover the life assisted scene, with the focus on the relevant high frequency words.

**Step 2.** Construct the seed text corpus $C_{seed}$ based on the cycle template of natural language generation method.

**Step 3.** The scale of the seed text corpus needs to be properly controlled. The corpus quality is used as a criterion to control the scale, and the scale adjustment can be returned to the step 2 iterative update.

**Step 4.** The step 3 result is judged manually, mainly due to the following considerations. The scale of the seed text corpus here is strictly controlled to a small extent, and we have relatively high quality requirements for the seed text corpus.

**Step 5.** Using the seed textcorpus as a training set, training based on the deep neural network model, the model $M_C$ is obtained.

**Step 6.** Based on the model $M_C$. Further inference, get the final meet-required task-oriented corpus $C_{DN}$.

**Step 7.** The task-oriented corpus $C_{DN}$ generated in step6 may not necessarily meet the requirements in some cases. For example, in the high diversity aspect, we need to go back to step5 and further iterate and optimize until it meets the requirements.

### 3.3 Constructing Seeds Text Corpus

Figure 1. seed lexicon

In this section, we introduce the method of constructing the seed Text Corpus.

**Step 1.** Limit the scope of life scenes

**Step 2.** According to the word frequency, extract common nouns, verbs, adjectives and quantifiers from the established scenes to form a seed lexicon, as shown in Figure 1.

**Step 3.** According to the rule structure of grammatical subject, predicate and object, the corresponding vocabulary is extracted from the seed lexicon to form a sentence.

**Step 4.** The statement is expanded to include appropriate quantifiers and adjectives before the corresponding nouns.

Example:

a1. I want to wear a coat

a2. I want to wear a long coat

a3. I want to wear a long coat with velvet

b1. He wants to drink water

b2. He wants to drink a glass of water

b3. He wants to drink a cup of warm water.

b4. He wants to drink a cup of warm water with sugar.

### 3.4 Constructing Text Corpus Based on Deep Neural Networks NLG Method

In this section, we designed the NLG model to implement non-fixed length input and generate non-fixed length output text. We need to construct the sequence-to-sequence structure. First implement embeddings for inputting non-fixed length text.

We use BERT[25][26] to implement the encoding function, where the BERT stands for means bidirectional encoder representations from transformers. In the process of using BERT, Due to the use of language models for transformer learning, empirical improvements have shown that unsupervised pre-training is an important integral part of many natural language processing systems. The BERT method suggests that a general language model may exist. Prior to this, when people solved NLP problems, they often customized specific models for specific problems. Howerver, BERT successfully solves a wide range of NLP tasks. Based on the above description, our structure is shown in Figure 2.

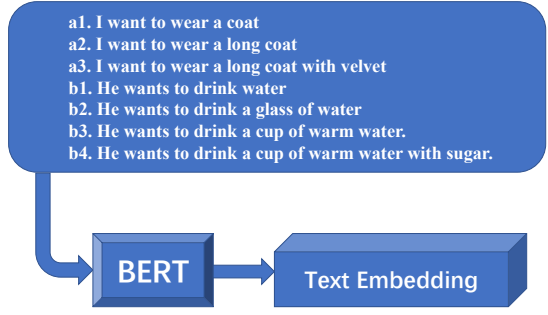

Figure 2. Embedding based on BERT

Based on the idea of these literature [20][27]-[29], We implement the natural language generation transformer model for specific text to text. Furthermore, we have adopted a method based on these documents[30]-[33]. The detailed process is described below. The basic structure of LSTM is as follows. Where $\varsigma_t$, $\zeta_t$, $\xi_t$ and $\psi_t$ represent input gate, forget gate, output gate and memory cell, respectively.

$$\varsigma_t = \sigma(\Xi_\varsigma \Re_{t-1} + \Im_\varsigma \vartheta_{x_t} + \beta_\varsigma)$$
$$\zeta_t = \sigma(\Xi_\zeta \Re_{t-1} + \Im_\zeta \vartheta_{x_t} + \beta_\zeta)$$
$$\xi_t = \sigma(\Xi_\xi \Re_{t-1} + \Im_\xi \vartheta_{x_t} + \beta_\xi)$$
$$\varphi_t = \tanh(\Xi_\varphi \Re_{t-1} + \Im_\varphi \vartheta_{x_t} + \beta_\varphi)$$
$$\psi_t = \zeta_t \odot \psi_{t-1} + \varsigma_t \odot \varphi_t$$
$$\Re_t = \xi_t \odot \tanh(\psi_t)$$

Where $\vartheta_{x_t}$ stand for the embedding of input. $\Xi$ and $\Im$ stand for corresponding matrix, and $\beta$ represent bias. They are all trainable parameters. $\Re_t$ stand for the hidden state of LSTM.
Utilizing bi-directional LSTM structure[31], we can extract text context information from both forward and backward directions. The hidden state can be expressed as follows.

$$\Re_t = \Re_{L_t} \oplus \Re_{R_t}$$

where $\Re_{L_t}$ and $\Re_{R_t}$ represent hidden state of backward direction and hidden state of forward direction, respectively. and $\oplus$ stands for the concat operation.
Output text sequence $\mathcal{L}_1, \mathcal{L}_2, \dots, \mathcal{L}_N$, has a defined generation rule[20][34]. The elements $\mathcal{L}_k, (k = 1,2,\cdots,N)$ are generated as follows.

$$p(\mathcal{L}_k|\mathcal{L}_1\mathcal{L}_2\cdots\mathcal{L}_{k-1}) = softmax(W(\sum_{e=1}^{M}((softmax(s_k \mathcal{W}\Re_e))\Re_e)))$$

Where $s_k$ has the following representation.

$$s_k = bi\text{-}LSTM((W^{out}(\mathcal{L}_{k-1}) \oplus \sum_{e=1}^{M}((softmax(s_{k-1}\mathcal{W}\Re_e))\Re_e)), s_{k-1})$$

Wherein, input text sequence $\vartheta_x = (\vartheta_{x_1}, \vartheta_{x_2}, \cdots, \vartheta_{x_M})$. where $W$, $\mathcal{W}$ and $W^{out}$ represent output matrix, attention matrix and output embedding matrices, respectively.

## 4. RESULTS AND ANALYSIS
Based on the methods presented in Sections 3.3 and 3.4, we combine the above two methods to construct a hybrid algorithm for task-oriented corpus.

Task-oriented hybrid model can learn to express structured input in an appropriate way. These advantages mainly come from the exquisite and elegant seeds text corpus design and deep neural networks natural language generation method. The hybrid model has two advantages including linguistic quality and high diversity.

In the field of SSR experiment, analysis results show that the performance of our hybrid model outperforms the pure method such as template-based natural language generation or neural natural language generation models.

## 5. ACKNOWLEDGMENTS
We are very grateful to DeepBlue Technology (Shanghai) Co., Ltd. and DeepBlue Academy of Sciences for their support.Thanks to the support of the Equipment pre-research project (No. 31511060502).

## 6. REFERENCES

[1] Bowman, M., Debray, S. K., and Peterson, L. L. 1993.A. B. Nassif, I. Shahin, I. Attili, M. Azzeh, and K. Shaalan. Speech Recognition Using Deep Neural Networks: A Systematic Review. IEEE Access, vol. 7, pp. 19143–19165, 2019.

[2] Z. Ling et al.. Deep Learning for Acoustic Modeling in Parametric Speech Generation: A systematic review of existing techniques and future trends. *IEEE Signal Process. Mag.*, vol. 32, no. 3, pp. 35–52, May 2015.

[3] G. Sharpe, V. Camoes Costa, W. Doubé, J. Sita, C. McCarthy, and P. Carding. Communication changes with laryngectomy and impact on quality of life: a review. *Qual. Life Res.*, Nov. 2018.

[4] J. Rajnoha and P. Pollák. ASR systems in Noisy Environment: Analysis and Solutions for Increasing Noise Robustness. vol. 20, no. 1, p. 11, 2011.

[5] B. Denby, T. Schultz, K. Honda, T. Hueber, J. M. Gilbert, and J. S. Brumberg. Silent speech interfaces. *Speech Commun.*, vol. 52, no. 4, pp. 270–287, Apr. 2010.

[6] M. J. Fagan, S. R. Ell, J. M. Gilbert, E. Sarrazin, and P. M. Chapman. Development of a (silent) speech recognition system for patients following laryngectomy. *Med. Eng. Phys.*, vol. 30, no. 4, pp. 419–425, May 2008.

[7] T. Hueber, E.-L. Benaroya, G. Chollet, B. Denby, G. Dreyfus, and M. Stone. Development of a silent speech interface driven by ultrasound and optical images of the tongue and lips. *Speech Commun.*, vol. 52, no. 4, pp. 288–300, Apr. 2010.

[8] R. Hofe et al.. Small-vocabulary speech recognition using a silent speech interface based on magnetic sensing. *Speech Commun.*, vol. 55, no. 1, pp. 22–32, Jan. 2013.

[9] G. N. Sonawane, A. N. Shewale, and G. G. Gujarathi. A Review on Modalities of EMG Based Speech Recognition

for Silent Speech Interface. p. 6, International Journal of Engineering Science and Computing, July 2016.

[10] Richard Li, Jason Wu, Thad Starner. 2019. TongueBoard: An Oral Interface for Subtle Input. In Augmented Human International Conference 2019 (AH2019), March 11–12, 2019, Reims, France. ACM, New York, NY, USA, Article 4, 9 pages. https://doi.org/10.1145/3311823.3311831

[11] H. Akbari, B. Khalighinejad, J. L. Herrero, A. D. Mehta, and N. Mesgarani. Towards reconstructing intelligible speech from the human auditory cortex. *Sci. Rep.*, vol. 9, no. 1, Dec. 2019.

[12] G. Krishna, C. Tran, J. Yu, and A. H. Tewfik. Speech Recognition with no speech or with noisy speech. Mar. 2019.

[13] C. H. Nguyen, G. K. Karavas, and P. Artemiadis. Inferring imagined speech using EEG signals: a new approach using Riemannian manifold features. *J. Neural Eng.*, vol. 15, no. 1, p. 016002, Feb. 2018.

[14] Dzmitry Bahdanau, Kyunghyun Cho, and Yoshua Bengio. Neural machine translation by jointly learning to align and translate. *arXiv preprint arXiv:1409.0473*, 2014.

[15] Abigail See, Peter J. Liu, and Christo- pher D. Manning. Get to the point: Summarization with pointer-generator networks. In *Proceedings of the 55th Annual Meeting of the Association for Computational Lin- guistics, ACL*, pages 1073–1083, 2017.

[16] Sebastian Gehrmann, Yuntian Deng, and Alexander M Rush. Bottom-up abstractive summarization. *arXiv preprint arXiv:1808.10792*, 2018.

[17] Wei Li, Xinyan Xiao, Yajuan Lyu, and Yuanzhuo Wang. Improving Neural Abstractive Document Summarization with Explicit Information Selection Modeling. In *EMNLP*, pages 1787–1796, 2018.

[18] Romain Paulus, Caiming Xiong, Richard Socher, and Palo Alto. A deep reinforced model for abstractive summarization. *ICLR*, pages 1–13, 2018.

[19] Tsung-Hsien Wen, Milica Gasic, Nikola Mrksic, Peihao Su, David Vandyke, and Steve J. Semantically Conditioned LSTM-based Natural Language Generation for Spoken Dialogue Systems. In Proceedings of the 2015 Conference on Empirical Methods in Natural Language Processing, EMNLP 2015, pages 1711–1721, Lisbon, Por- tugal.

[20] Glorianna Jagfeld, Sabrina Jenne, Ngoc Thang Vu. Sequence-to-Sequence Models for Data-to-Text Natural Language Generation: Word- vs. Character-based Processing and Output Diversity. *arXiv preprint arXiv:* 1810.04864, 2018.

[21] Chris Alberti, Kenton Lee, Michael Collins. A BERT Baseline for the Natural Questions. arXiv preprint arXiv:1901.08634. 2019

[22] Alex Wang, Kyunghyun Cho. BERT has a Mouth, and It Must Speak: BERT as a Markov Random Field Language Model. arXiv preprint arXiv:1902.04094 .2019

[23] Qian Chen, Zhu Zhuo, Wen Wang. BERT for Joint Intent Classification and Slot Filling. arXiv preprint arXiv:1902.10909. 2019

[24] Chanwoo Jeong, Sion Jang, Hyuna Shin, Eunjeong Park, Sungchul Choi. A Context-Aware Citation Recommendation Model with BERT and Graph Convolutional Networks. arXiv preprint arXiv:1903.06464. 2019

[25] Jacob Devlin, Ming-Wei Chang, Kenton Lee, Kristina Toutanova. BERT: Pre-training of Deep Bidirectional Transformers for Language Understanding. arXiv preprint arXiv:1810.04805. 2018

[26] Haoyu Zhang, Yeyun Gong, Yu Yan, Nan Duan, Jianjun Xu, Ji Wang , Ming Gong and Ming Zhou. Pretraining-Based Natural Language Generation for Text Summarization. *arXiv preprint arXiv:* 1902.09243,2019.

[27] Kyunghyun Cho, Bart Van Merrie ̈nboer, C ̧alar Gu ̈lc ̧ehre, Dzmitry Bahdanau, Fethi Bougares, Hol- ger Schwenk, and Yoshua Bengio. Learning phrase representations using rnn encoder–decoder for statistical machine translation. In Proceedings of the 2014 Conference on Empirical Methods in Nat- ural Language Processing (EMNLP), pages 1724–1734, Doha, Qatar.

[28] Ashish Vaswani, Noam Shazeer, Niki Parmar, Jakob Uszkoreit, Llion Jones, Aidan N. Gomez, Lukasz Kaiser, Illia Polosukhin. Attention Is All You Need. arXiv:1706.03762

[29] Zhiheng Huang, Wei Xu, Kai Yu. Bidirectional LSTM-CRF Models for Sequence Tagging. arXiv:1508.01991

[30] Qi Zhang, Xiaoyu Liu, Jinlan Fu. Neural Networks Incorporating Dictionaries for Chinese Word Segmentation. Association for the Advancement of Artificial Intelligence, 2018.

[31] Graves, A., and Schmidhuber, J. 2005. Framewise phoneme classification with bidirectional lstm networks. In Neural Networks, 2005. IJCNN'05. Proceedings. 2005 IEEE International Joint Conference on, volume 4, 2047–2052.

[32] David Ha, Andrew Dai, Quoc V. Le. HyperNetworks. arXiv:1609.09106. 2016

[33] Hochreiter, S., Schmidhuber, J. Long short-term memory. Neural computation 9(8):1735–1780. 1997.

[34] Thang Luong, Hieu Pham, and Christopher Manning. Effective approaches to attention-based neural machine translation. In Proceedings of the 2015 Conference on Empirical Methods in Natural Lan- guage Processing, EMNLP 2015, pages 1412–1421, Lisbon, Portugal.